\documentclass[runningheads]{llncs}
\usepackage[T1]{fontenc}

\usepackage{graphicx}
\usepackage{soul,xcolor}
\usepackage{amsmath}
\usepackage{amssymb}
\usepackage{subfigure}
\usepackage{hyperref}
\usepackage{csquotes}

\begin{document}
\title{BBOB Instance Analysis}
\subtitle{Landscape Properties and Algorithm Performance across Problem Instances}
\titlerunning{BBOB Instance Analysis}
\author{Fu Xing Long\inst{1}\orcidID{0000-0003-4550-5777} \and
Diederick Vermetten\inst{2}\orcidID{0000-0003-3040-7162} \and
Bas van Stein\inst{2}\orcidID{0000-0002-0013-7969} \and \\
Anna V. Kononova\inst{2}\orcidID{0000-0002-4138-7024}}
\authorrunning{F. Long et al.}

\institute{BMW Group,
Knorrstra{\ss}e 147, D-80788 Munich, Germany\\
\email{fu-xing.long@bmw.de}\\ \and
LIACS, Leiden University, Niels Bohrweg 1, NL-2333 Leiden, Netherlands\\
\email{\{d.l.vermetten,b.van.stein,a.kononova\}@liacs.leidenuniv.nl}}
\maketitle              % typeset the header of the contribution
\begin{abstract}
Benchmarking is a key aspect of research into optimization algorithms, and as such the way in which the most popular benchmark suites are designed implicitly guides some parts of algorithm design. 
One of these suites is the black-box optimization benchmarking (BBOB) suite of $24$ single-objective noiseless functions, which has been a standard for over a decade. 
Within this problem suite, different instances of a single problem can be created, which is beneficial for testing the stability and invariance of algorithms under transformations. 
In this paper, we investigate the BBOB instance creation protocol by considering a set of $500$ instances for each BBOB problem.
Using exploratory landscape analysis, we show that the distribution of landscape features across BBOB instances is highly diverse for a large set of problems.
In addition, we run a set of eight algorithms across these $500$ instances, and investigate for which cases statistically significant differences in performance occur. 
We argue that, while the transformations applied in BBOB instances do indeed seem to preserve the high-level properties of the functions, their difference in practice should not be overlooked, particularly when treating the problems as box-constrained instead of unconstrained.

\keywords{Exploratory landscape analysis \and Black-box optimization \and Benchmarking \and 
Single-objective optimization \and Instance spaces.}
\end{abstract}

\section{Introduction} \label{sec:intro}
Solving black-box optimization (BBO) problems can be extremely challenging, even with domain knowledge and experience. Due to the fact that no analytical form is available, derivative information is lacking and numerical approximation of the derivatives is costly \cite{bbo_audet2017}. 
The task becomes particularly tedious and cumbersome when it comes to real-world BBO problems with expensive function evaluation, e.g., crash-worthiness optimization in automotive industry that requires simulation runs \cite{ela_long2022}. 
Since developing and testing algorithms on these real-world problems is prohibitively expensive, benchmarking on artificial test problems (of similar problem classes) becomes necessary to gain an understanding of the algorithm, which can then hopefully be transferred to the original scenario. 

Since benchmarking on artificial test problems is generally done to gain insight into the behavior of the algorithm under known conditions, various benchmarking suites have been developed, where different global function properties are represented, such as multi-modality, different types of global structure and separability. 
One of the most well-known suites for single-objective, noiseless and continuous optimization is often called \textit{the} BBOB suite~\cite{bbob_hansen2009_noiseless}, originally proposed as part of the comparing continuous optimizers (COCO) environment~\cite{coco_hansen2021}.
A key feature of BBOB is the fact that the functions can be scaled to arbitrary dimensionality, and that multiple different versions (\textit{instances}) of the same function can be created by applying some transformation methods to the underlying function, which are said to preserve the main properties of the function. 

For instance, different instances have been considered to enable comparisons between stochastic and deterministic optimization algorithms~\cite{povsik2009bbob}, since using a different instance can in some way be considered as changing the initialization of the deterministic algorithm. 
It also enables an algorithm designer to test for some invariance properties, particularly with regard to scaling of the objective values, and rotation of the search space~\cite{bbob_hansen2009_noiseless}. 
Recently, instances have also been used in a more Machine Learning (ML) based context, e.g., methods for algorithm selection are trained and tested based on different sets of instances~\cite{traj_dynas}. 

While creation of different instances of the same function has been very useful to many benchmarking setups, \textit{the underlying assumption that the function properties are preserved is a rather strong one}. 
For a simple sphere function, the impact of moving the optimum throughout the space can be reasoned about relatively easily, but the impact of the more involved transformation methods on more complex functions is challenging to be quantified directly. 
In addition, the fact that black box optimization problems are in practice often considered to be box-constrained~\cite{bartz2020benchmarking}, while BBOB was originally designed based on unconstrained function definitions~\cite{coco_hansen2021}, introduces the possibility that some transformations might change key aspects of the function. 

In order to analyze the resulting low-level properties of optimization problems, various features of the landscape can be computed. 
This falls under the field of Exploratory Landscape Analysis (ELA)~\cite{ela_mersmann2010}. While some analysis into the ELA features across instances of BBOB problems has been previously performed~\cite{ela2_munoz2022}, we extend the scope of our analysis to include a much wider range of instances. 
In addition, we consider several other low-level features, such as the location of the global optima, to develop an extensive understanding of the way in which instances might differ. 
Since we deal with the box-constrained version of the BBOB problems, we also investigate the performance of a set of algorithm, in order to verify that these algorithms perform similar on different instances of a function -- this extends approach taken in~\cite{cmaes_vermetten2022}.
In particular, we aim to answer the following research questions:
\begin{enumerate}
    \item How well are the problem characteristics of a particular BBOB function preserved across different problem instances?
    \item How representative is the first BBOB problem instance, or the first few instances?
    \item Is there any significant difference in algorithm performances across different problem instances of the same BBOB function? 
\end{enumerate}

The remainder of this paper is structured as follows:
Section \ref{sec:sota} briefly introduces the BBOB suite and ELA method.
This is followed by an overview on the experimental setup in Section \ref{sec:setup}.
Research results concerning landscape characteristics are discussed in Section \ref{sec:result_ela}, algorithm performances in Section \ref{sec:result_algo} and properties of instances in Section \ref{sec:result_prop}.
Lastly, conclusions and future works are provided in Section \ref{sec:conclusion}.

\section{Related Work} \label{sec:sota}

Without loss of generality, a continuous BBO problem can be (typically) defined as the minimization of an objective function $f \colon \mathcal{X} \rightarrow \mathcal{Y}$, where $\mathcal{X} \subseteq \mathbb{R}^d$ is the search space, $\mathcal{Y} \subseteq \mathbb{R}$ is the objective space and $d$ is the dimensionality.
Over the years, various state-of-the-art derivative-free heuristic optimization algorithms have been developed to handle these BBO problems. Since analysis of these methods for general black-box functions is infeasible, comparisons between them rely on benchmark suites which cover different classes of functions. Such comparisons are then expected to be performed whenever a new algorithm or algorithmic variant is proposed, which naturally leads to these same benchmark problems playing a major role during algorithm development. As such, the exact construction of the commonly used benchmark functions to some extend guides the direction of algorithm development. Gaining a more thorough understanding of these benchmark suites would then allow us to uncover potential biases in the types of problems algorithm are being tested on and, thus, investigate the generalisability of results. 

\subsection{COCO and the BBOB Benchmark Suite}
COCO~\cite{coco_hansen2021}, as one of the most established tools for benchmarking optimization heuristics, enables a fair comparison between algorithms by recording detailed performance statistics, which can be processed and compared to a wide set of publicly accessible data\footnote{\url{https://numbbo.github.io/data-archive/bbob/}} from other state-of-the-art optimization algorithms -- this repository is constantly expanded by users, in part through the yearly BBOB workshops. 
Within COCO, the most used suite of functions is the BBOB suite for single-objective, noiseless, continuous optimization (we refer to this suite as BBOB throughout this paper). 
This suite contains 24 functions, which can be separated into five core classes based on their global properties. 
While the suite is originally intended to be used for unconstrained optimization, in practice however, black box optimization functions like this are often considered to be box-constrained~\cite{bartz2020benchmarking}, in the case of BBOB with domain $[-5,5]^d$. 
Such distinction is however not reflected in the aforementioned repository. 

For each BBOB function, arbitrarily many problem instances can be generated by applying transformations to both the search space and the objective values~\cite{bbob_hansen2009_noiseless} -- such mechanism is implemented internally in BBOB and controlled via a unique identifier (also known as IID) which defines the applied transformations (e.g. rotation matrices). 
For most functions, the search space transformation are made up of rotations and translations (moving the optimum, usually uniformly in $[-4,4]^d$). 
Since the objective values can also be transformed, the performance measures used generally are relative to the global optimum value to allow for comparison of performance between instances, typically in logarithmic scale. 

While the instance generation is certainly useful for many applications, it has not been without critique. 
In particular, the stability of low-level features under the used transformations might not be guaranteed~\cite{ela2_munoz2022}. In this work, we focus on identifying these potential differences. 

\subsection{Exploratory Landscape Analysis} \label{sec:ela}
In general, ELA provides an automated approach to estimate the complexity of an optimization problem, by capturing its topology or landscape characteristics.
More precisely, the high-level landscape characteristics of optimization problems, such as multi-modality, global structure and separability, are numerically quantified through six classes of expertly designed low-level features, namely $y$-distribution, level set, meta-model, local search, curvature and convexity \cite{ela_mersmann2010,ela_mersmann2011}.
These landscape characteristics, also known as ELA features, can be cheaply computed based on a Design of Experiments (DoE), consisting of some samples and their corresponding objective values.

In recent years, ELA has gained increasing attention in the landscape-aware algorithm selection problem (ASP) tasks, where the correlation between landscape characteristics and optimization algorithm performances has been intensively researched.
In fact, previous works have revealed that ELA features are indeed informative in explaining algorithm behaviors and can be exploited to reliably predict algorithm performances, e.g., using ML approach \cite{ela_bischl2012,ela_kerschke2019,ela_jankovic2020}.
Apart from ASP tasks, ELA has shown promising potential in other application domains, for instance, classification of the BBOB functions \cite{ela_renau2021} and instance space analysis of different benchmark problem sets \cite{ela_skvorc2020}.
While we are fully aware that the ELA features are highly sensitive to sample size \cite{ela2_munoz2022} and sampling strategy \cite{ela_renau2020,ela2_skvorc2021}, these aspects are beyond the scope of this research.

\section{Experimental Setup and Reproducibility}\label{sec:setup}
In this work, we consider the first 500 instances of the BBOB test suite of $5d$ and $20d$ and access them using the \texttt{IOHexperimenter}~\cite{iohexperimenter} package.
Throughout this work, all statistical tests are available in the package \texttt{scipy} \cite{scipy_virtanen2020} and we consider a confidence level of $99\%$, i.e. the null hypothesis is rejected, if the p-value is smaller than $0.01$. 
To ensure reproducibility, we have uploaded our experiments to a Figshare repository~\cite{repository_reproducibility}. 
In addition, figures which could not be included due to space-constraints have been uploaded to a separate Figshare repository~\cite{repository_reproducibility}. 

\section{Instance Similarity using ELA} \label{sec:result_ela}
\paragraph{Setup.} 
We first focus on analyzing the problem characteristics of different BBOB problem instances based on the ELA approach.
For each BBOB instance, we generate $100$ sets of DoE data with $1\,000$ samples each using the Latin Hypercube sampling (LHS) method (so the DoEs are identical for all instances), in order to obtain the ELA feature distribution.
We consider a total of $68$ ELA features that can be computed without additional sampling, using the package \texttt{flacco} \cite{flacco_kerschke2019,github_kerschke2019} and the pipeline developed in \cite{ela_long2022}. 
Three of the ELA features, which resulted in the same value across all instances, are deemed not informative and hence dropped out; this means that a final set of $65$ ELA features is being considered here.
 
\begin{figure}[!bt]
 \centering
 \includegraphics[width=\linewidth,trim=0mm 57mm 0mm 0mm,clip]{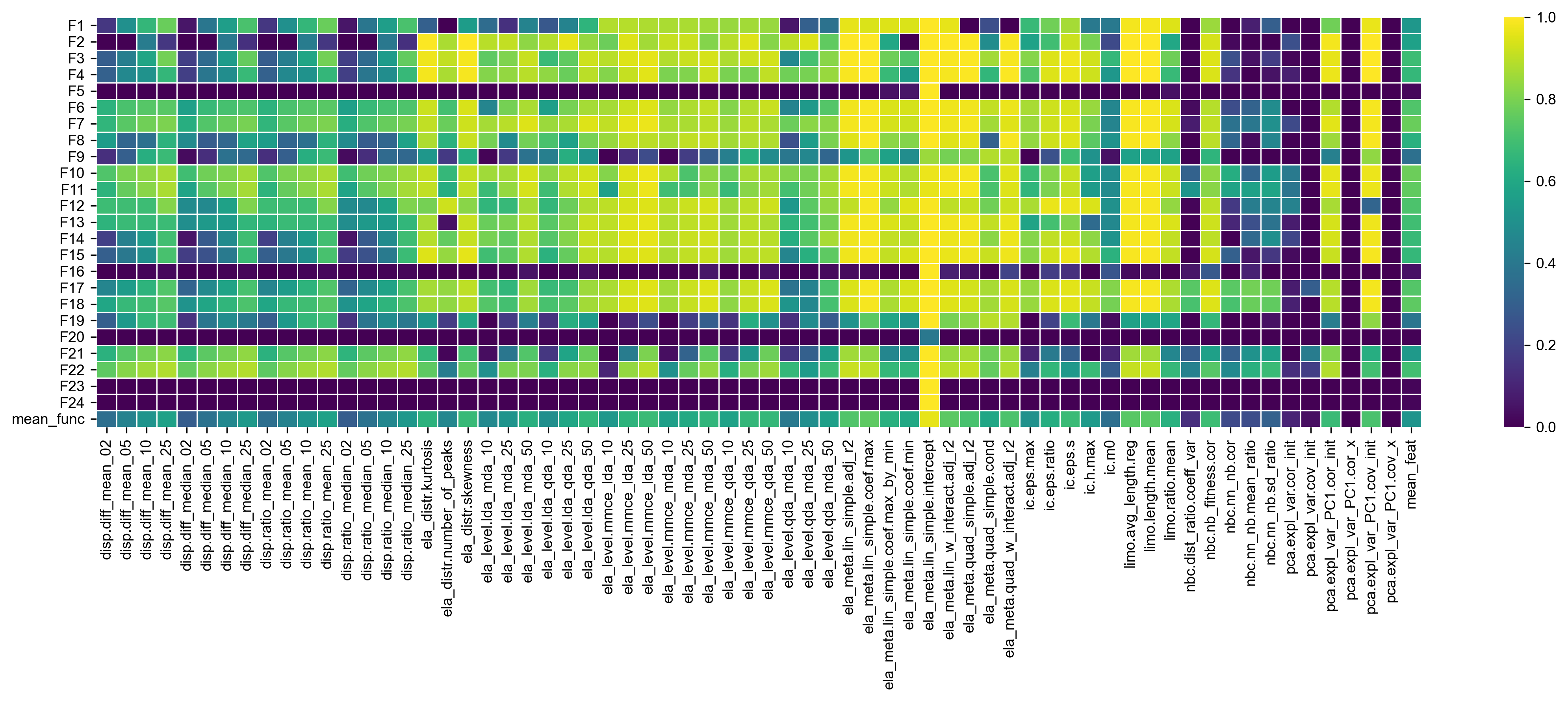}
 \includegraphics[width=\linewidth,trim=0mm 2mm 0mm 0mm,clip]{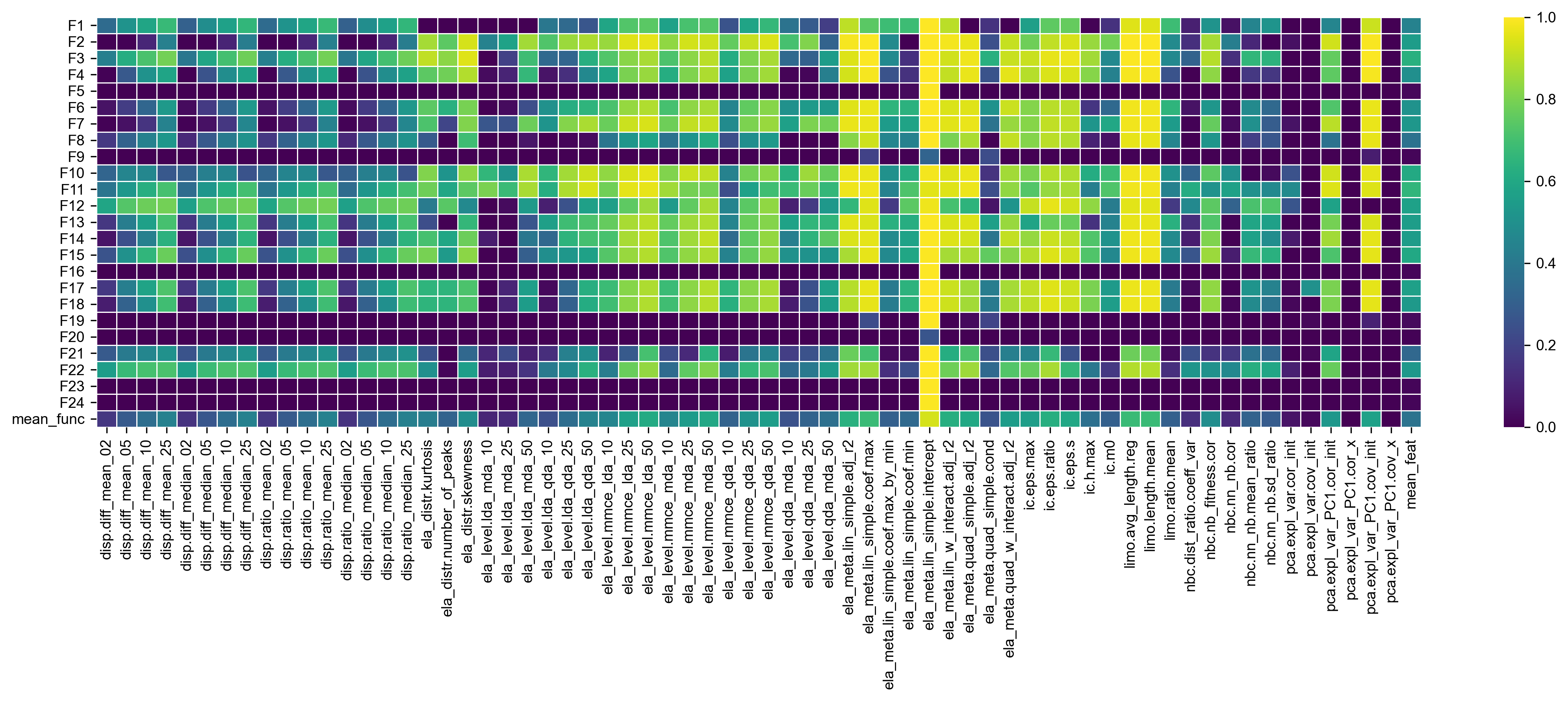}
 \caption{Average rejection rate of null hypothesis \textit{distribution of ELA feature between instances is similar}, aggregated over 500 BBOB problem instances in $5d$ (top) and $20d$ (bottom). A lighter color represents higher rejection rate. An extra row (bottom) for the mean over all BBOB functions and an extra column (right) for the mean over all ELA features in each heatmap.}
 \label{fig:ks_agg_ela}
\end{figure}

\paragraph{Comparing Distributions.}
To investigate how comparable the characteristics of different problem instances are, we carry out the (pairwise) two-sample Kol\-mo\-go\-rov-Smirnov (KS) test \cite{kstest_massey1951}, with the null hypothesis that the ELA distribution is similar in both (compared) problem instances. 
This results in $\frac{500\cdot499}{2}=124\,750$ comparison pairs per ELA feature.
To account for multiple comparisons, we apply the Benjamini-Hochberg (BH) method \cite{fdr_benjamini199}.
To get an overview of differences for a particular ELA feature of each BBOB function, we compute the average rejection rate of the aforementioned null hypothesis of the KS test by aggrega\-ting all problem instances (i.e. number of rejections divided by total number of tests).
In other words, it shows the fraction of tests which rejected each combination of ELA feature and BBOB function, as shown in Figure~\ref{fig:ks_agg_ela}.

On the $5d$ problems, we notice that some features clearly differ between instances, in particular the \verb|ela_meta.lin_model.intercept|. 
However, this does not necessarily indicate that all instances should indeed be considered to be different since, as illustrated in~\cite{ela2_skvorc2021}, some features including this linear model intercept are not invariant to scaling of the objective function. 
For some other features, such as those related to the principal component analysis (PCA), we notice that barely any test rejections are found. 
This is largely explained by considering that this feature-set is built primarily on the the samples in $\mathcal{X}$, which are identical between instances (same 100 seeds are used in the calculations for each instance ).
While the objective values in $\mathcal{Y}$ still have an influence on some of the PCA-features, their impact is relatively minor. 
For the remaining sets of features, we see some commonalities on a per-function basis. 
Functions F5 (linear slope), F16 (Weierstrass), F23 (Katsuura) and F24 (Lunacek bi-Rastrigin) show no difference between instances.

It is worthwhile to point out that even for a simple function as F1 (sphere), many features differ between instances. 
Since translation is the only transformation applied in F1~\cite{bbob_hansen2009_noiseless}, which (uniformly at random) moves the optimum to a point within $[-4,4]^d$, 
it is clear that the high-level function properties are preserved. If the problem is considered unconstrained, this transformation would indeed be a trivial change to the problem. However, since for ELA analysis, we are required to draw samples in a bounded domain, we have to consider the problems as box-constrained, and thus moving the function can have a significant impact on the low-level landscape features. This might explain why many ELA-features differ greatly across instances on the sphere function. We elaborate on this further in Section~\ref{sec:result_prop}.

On the other hand, the same overall patterns can be seen in $20d$ as on $5d$, albeit with a reduced magnitude.
Moreover, functions F9 (Rosenbrock), F19 (Composite Griewank-Rosenbrock) and F20 (Schwefel) now barely show any statistical difference between instances.

\begin{figure}[!ht]
 \centering
 \includegraphics[width=.525\linewidth,trim=0mm 0mm 0mm 0mm,clip]{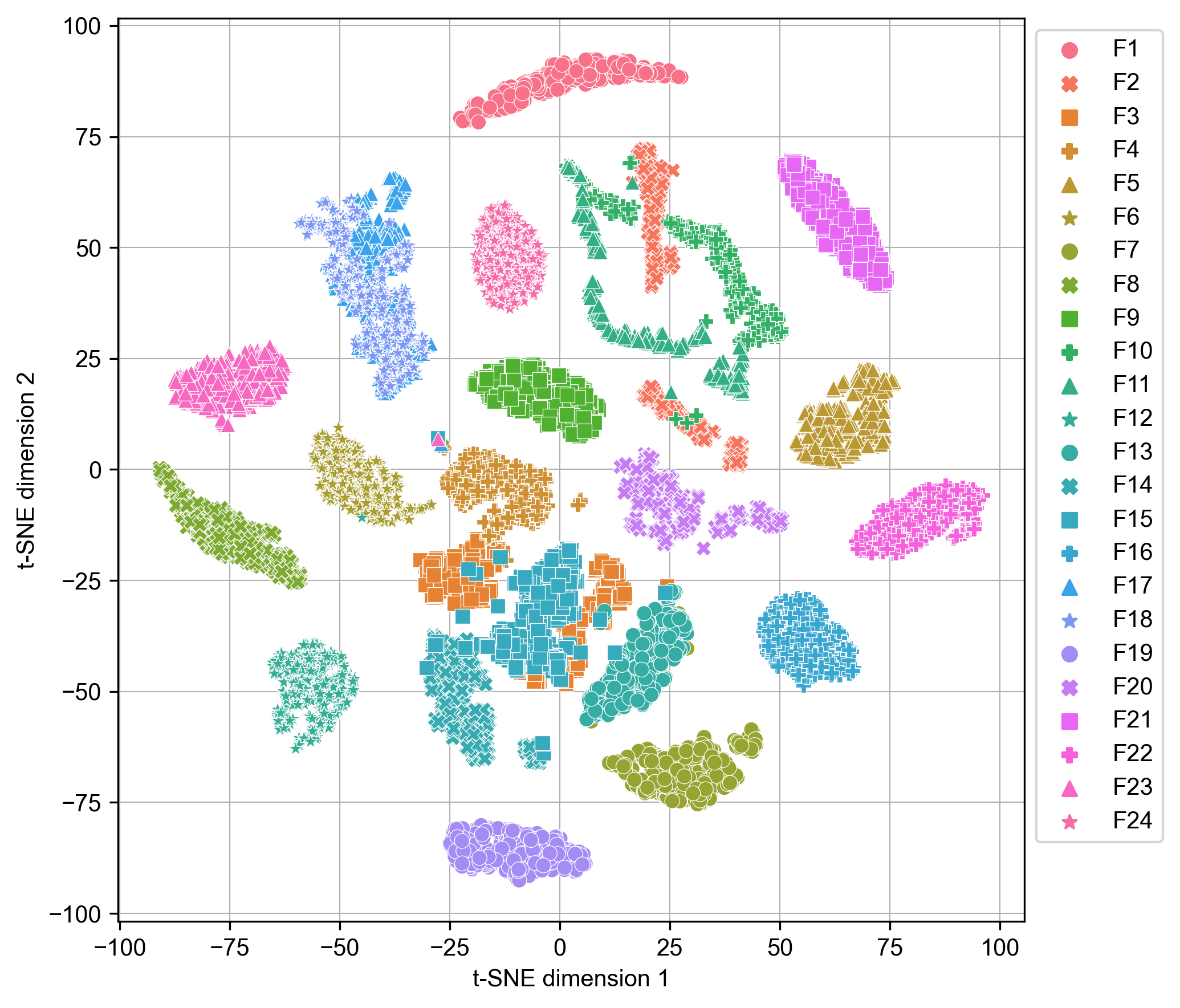}
 \includegraphics[width=.46\linewidth,trim=0mm 0mm 21.5mm 0mm,clip]{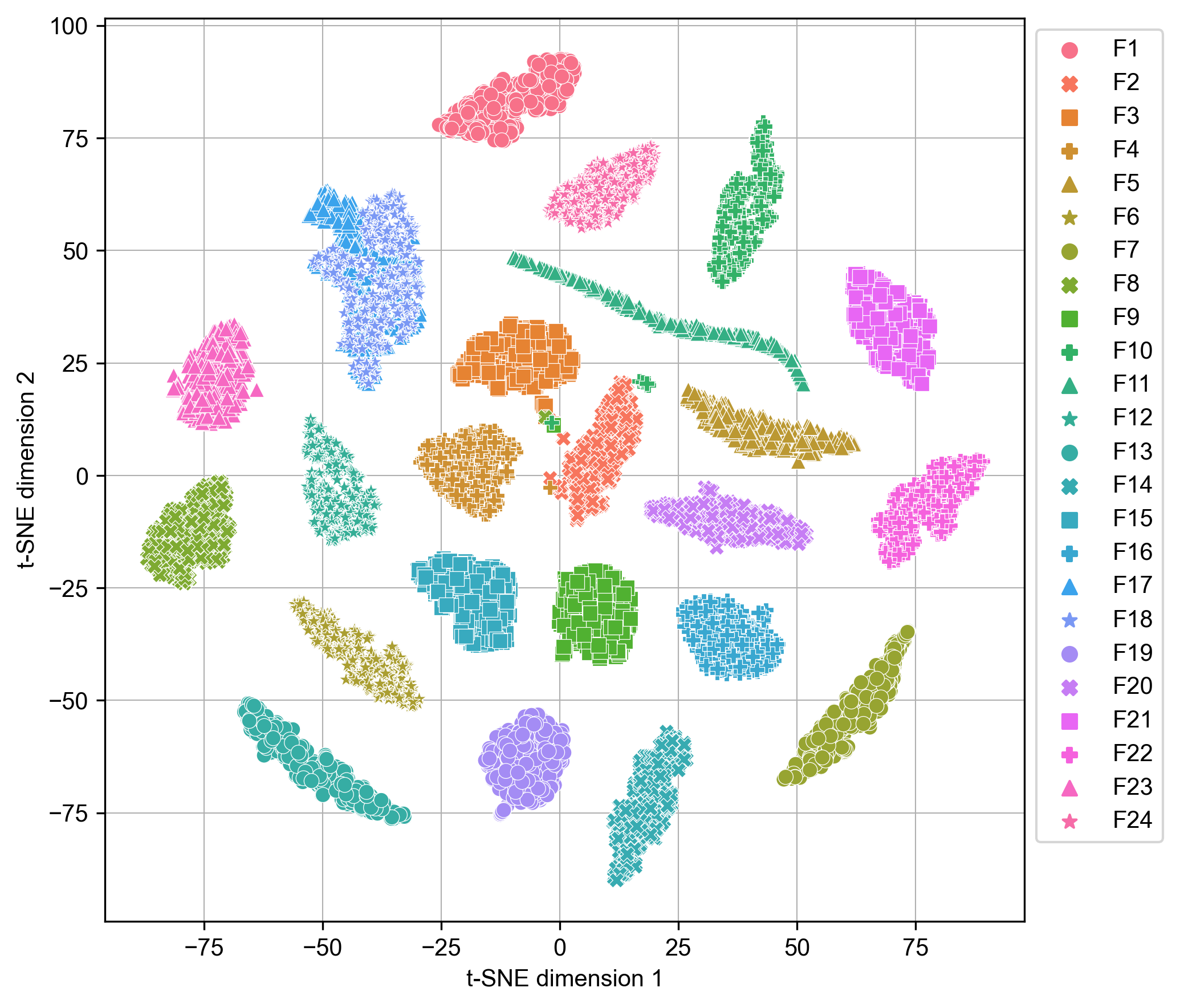}
 \caption{Projection of high-dimensional ELA feature space (altogether $64$ features, without \texttt{ela\_meta.lin\_simple.intercept}) onto a $2d$ visualization for the BBOB problems of $5d$ (left) and $20d$ (right) using t-SNE approach with default settings.}
 \label{fig:ela_space}
\end{figure}

\paragraph{Dimensionality Reduction.} 
In addition to the statistical comparison approach, we visualize the ELA features in a $2d$ space using the t-distributed stochastic neighbor embedding (t-SNE) approach \cite{tsne_maaten2008}, as shown in Figure~\ref{fig:ela_space} for features standardized beforehand by removing mean and scaling to unit variance.
It is clear that most instances of each problem are tightly clustered together. 
Nonetheless, there are outliers, where several instances of a function are spread throughout the projected space, indicating that these instances might be less similar. 
This is particularly noticeable in $5d$, where several functions are somewhat spread throughout the reduced space.
In $20d$, function clusters appear much more stable, matching the conclusion from the differences with regard to dimensionality in Figure~\ref{fig:ks_agg_ela}. 
It is worthwhile to note that differences between BBOB functions are indeed \textit{easier to be detected} in higher dimensions using ELA features, as shown in previous work~\cite{ela_renau2021}, which matches the more well-defined problem clusters we see in Figure~\ref{fig:ela_space}. 

\paragraph{Representativeness of instances 1--5.}
Many studies involving the BBOB suite make use of a small subset of available instances. 
For the BBOB workshops, the exact instances used have varied over time, but generally consist of 5 to 15 unique IIDs. 
Outside of the workshops, a common approach seems to be considering the first five instances only, and perform multiple repetitions on those. 
If these instances are not representative of the overall space of instances, such choice could potentially have an impact on results. 
Therefore we aim to verify the representativeness of these five instances. 
This can be achieved through considering the pairwise tests done for Figure~\ref{fig:ks_agg_ela}, but instead of aggregating the rejections on a per-feature level, we can do it per-function. 
If for an instance, the fraction of test rejections against other instances is high, while the overall fraction of pairwise test rejections is low, we can consider this instance to be an outlier, and thus non-representative.

\begin{figure}[!ht]
 \centering
 \includegraphics[width=\linewidth,trim=2mm 7mm 2mm 2mm,clip]{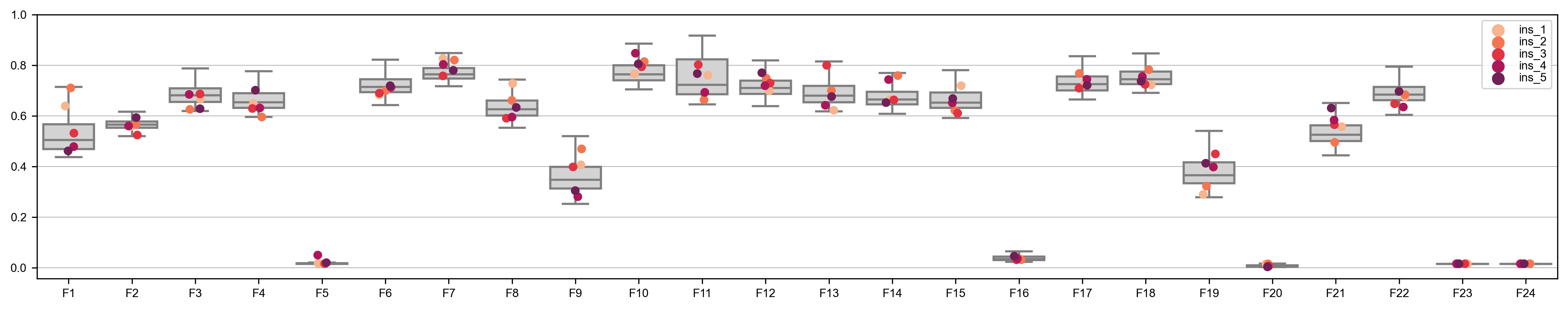}
 \includegraphics[width=\linewidth,trim=2mm 3mm 2mm 2mm,clip]{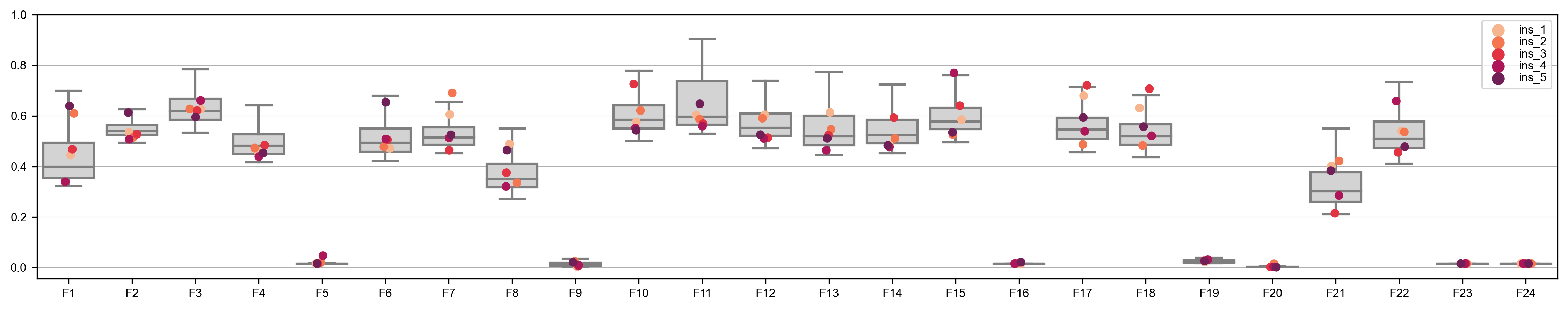}
\caption{Mean fraction of rejections of pairwise tests from one instance to each of the remaining ones, aggregated over all ELA features from Figure~\ref{fig:ks_agg_ela}. The first five instances of each function are highlighted, while the boxplots show the distribution for all the 500 instances for $5d$ (top) and $20d$ (bottom) functions.}
 \label{fig:ks_agg_repre}
\end{figure}

In Figure~\ref{fig:ks_agg_repre}, we visualize the average fraction of rejections (across features) in a boxplot, and highlight the rejection rate for each of the first five instances. 
We can see that there is no obvious case in which these five instances are all outliers. 
While some individual instances might have slightly more or slightly less rejections than the remaining instances, we could not conclude that the choice of selecting these five would be any worse than a different set of instances from the same function. 

\begin{figure}[!tb]
\centering
\includegraphics[height=.205\linewidth,trim=2mm 2mm 2mm 2mm,clip]{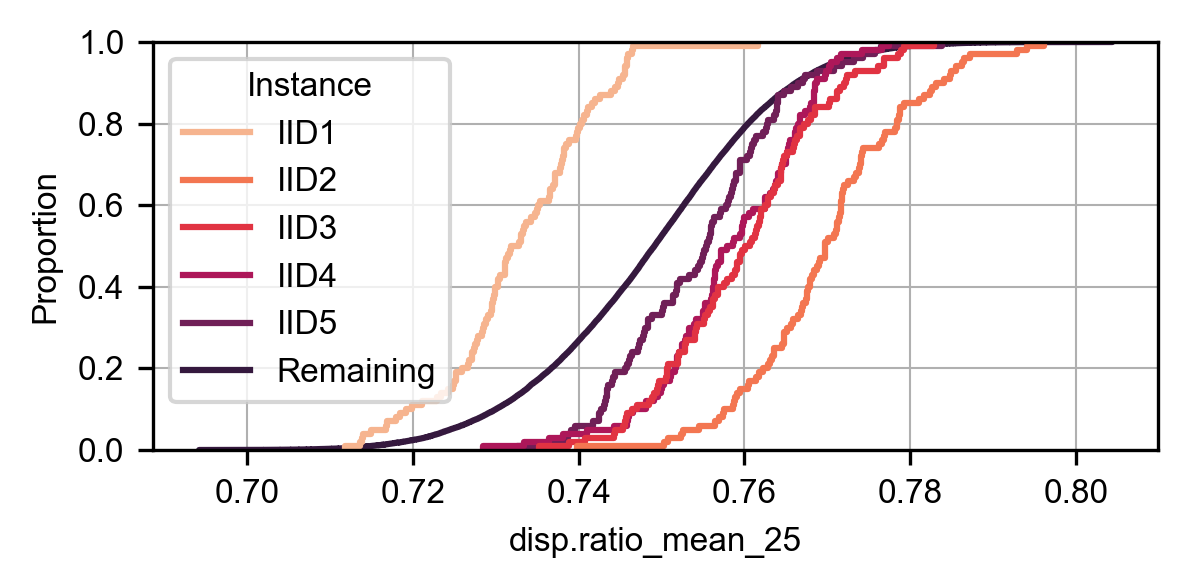}
\includegraphics[height=.205\linewidth,trim=12mm 2mm 2mm 2mm,clip]{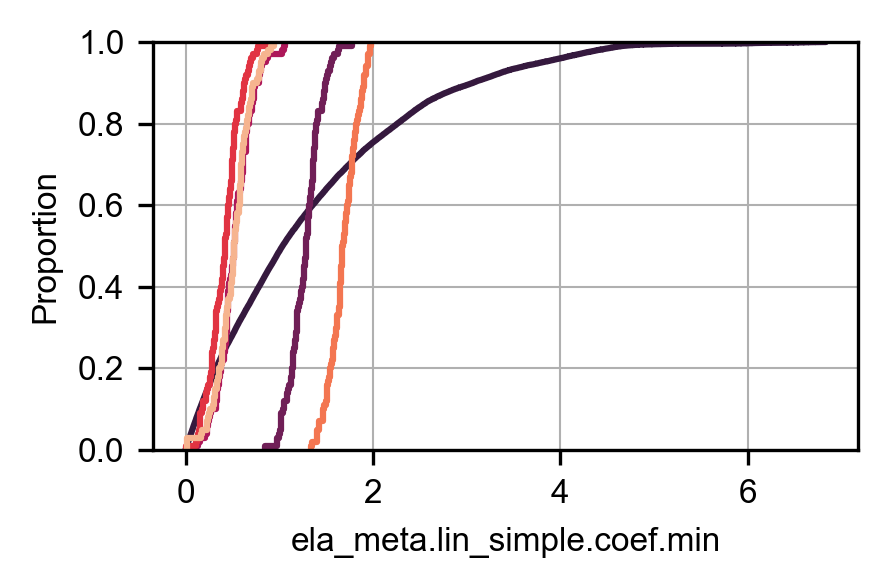}
\includegraphics[height=.205\linewidth,trim=12mm 2mm 2mm 2mm,clip]{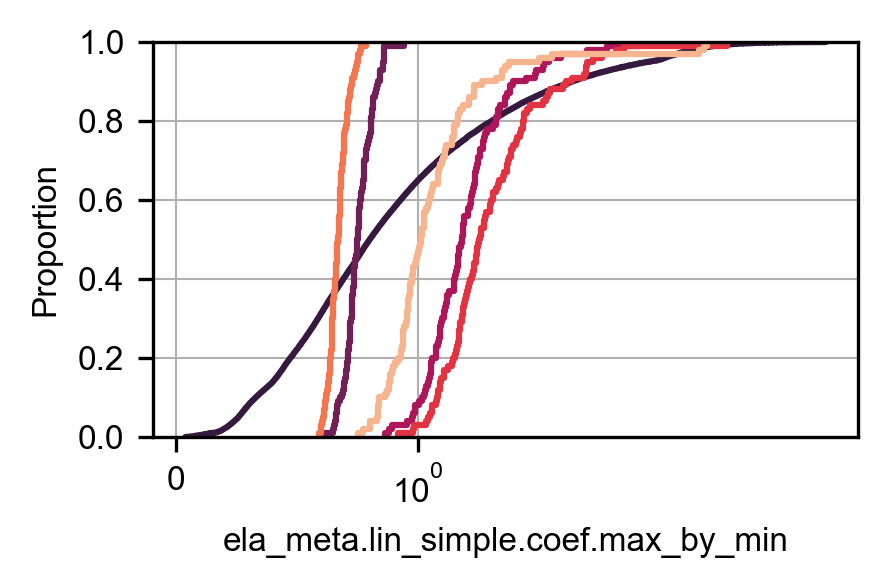}
\caption{Examples of ECDF curves of normally (left) and non-normally distributed (middle, right) ELA features on F1 of $5d$ for instances 1--5 against all remaining instances.} \label{fig:dist_ela}
\end{figure}

Apart from considering the overall representativeness from the aggregation of all features, we can also look into the individual ELA features in more detail. 
Because of space limitations, we show only three features for F1 in Figure~\ref{fig:dist_ela}. 
These empirical cumulative distribution curves (ECDF) show the differences in distribution of instances 1--5 against the remaining instances on three features, where the pairwise statistical test showed a large number of rejections. For this figure, we see that the differences between instances can be relatively large, but we see no evidence to conclude that the first five instances would be less representative than any other set. 

In addition to the represetativeness, we can also observe some interesting differences between features in terms of their distributions. While some features are seemingly normally distributed, we note that this is not the case for all features. In fact, we perform normality tests on each distribution, which highlight that several features, including distribution and most meta-model features, are often \textit{non-Gaussian} -- such figures are omitted here due to space limitation and can be found in supplementary material on Figshare~\cite{repository_reproducibility}. 

\section{Algorithm Performance across Instances}\label{sec:result_algo}
We now analyze the optimization algorithm performances across different BBOB problem instances.
Here, we consider single-objective unconstrained continuous optimization with the following eight derivative-free optimization algorithms available in Nevergrad \cite{ng_rapin2018} (all with default settings as set by Nevergrad): DiagonalCMA (a variant of covariance matrix adaptation evolution strategy (CMA-ES)), differential evolution (DE), estimation of multivariate normal algorithm (EMNA), NGOpt14, particle swarm optimization (PSO), random search (RS), constrained optimization by linear approximation with random restart (RCobyla) and simultaneous perturbation stochastic approximation (SPSA).
We run each algorithm on each of the $500$ instances of the $5d$ BBOB problems, 50 independent runs each, resulting in a total of $4.8$ million (=$8$ $\times$ $24$ $\times$ $500$ $\times$ $50$) algorithm runs, each run having a budget of $10\,000$ function evaluations.

\begin{figure}[!b]
\centering
\includegraphics[width=.5\linewidth,trim=2mm 18mm 17mm 2mm,clip]{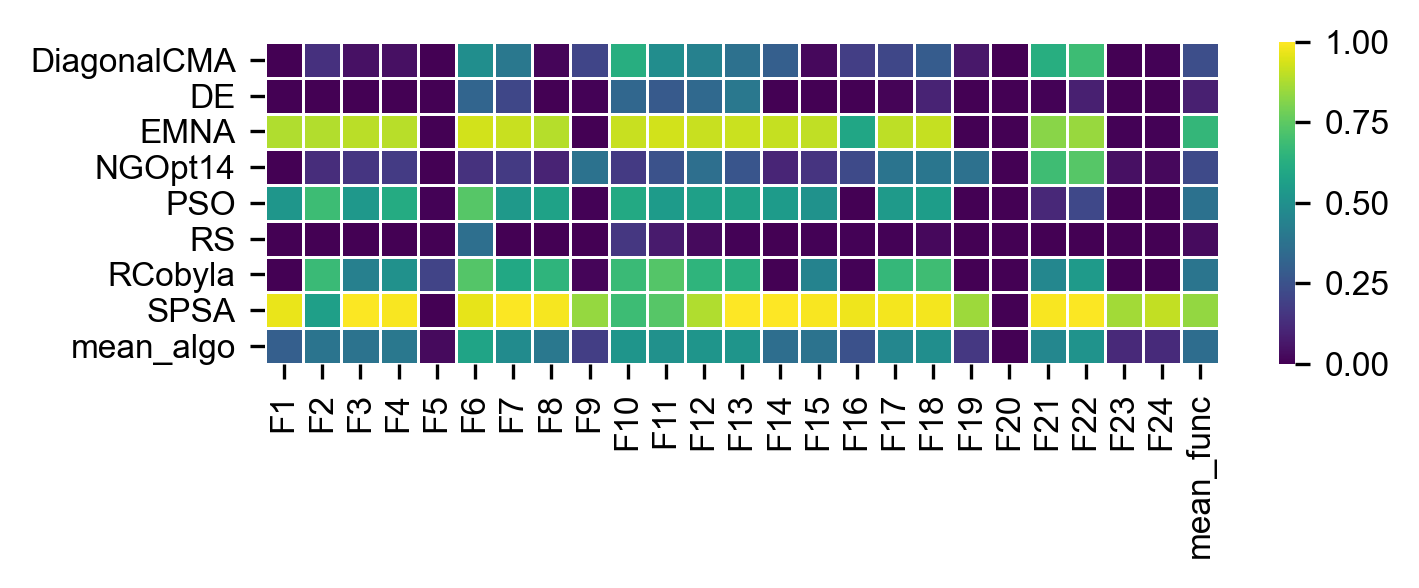}
\includegraphics[width=.49\linewidth,trim=20mm 18mm 2mm 2mm,clip]{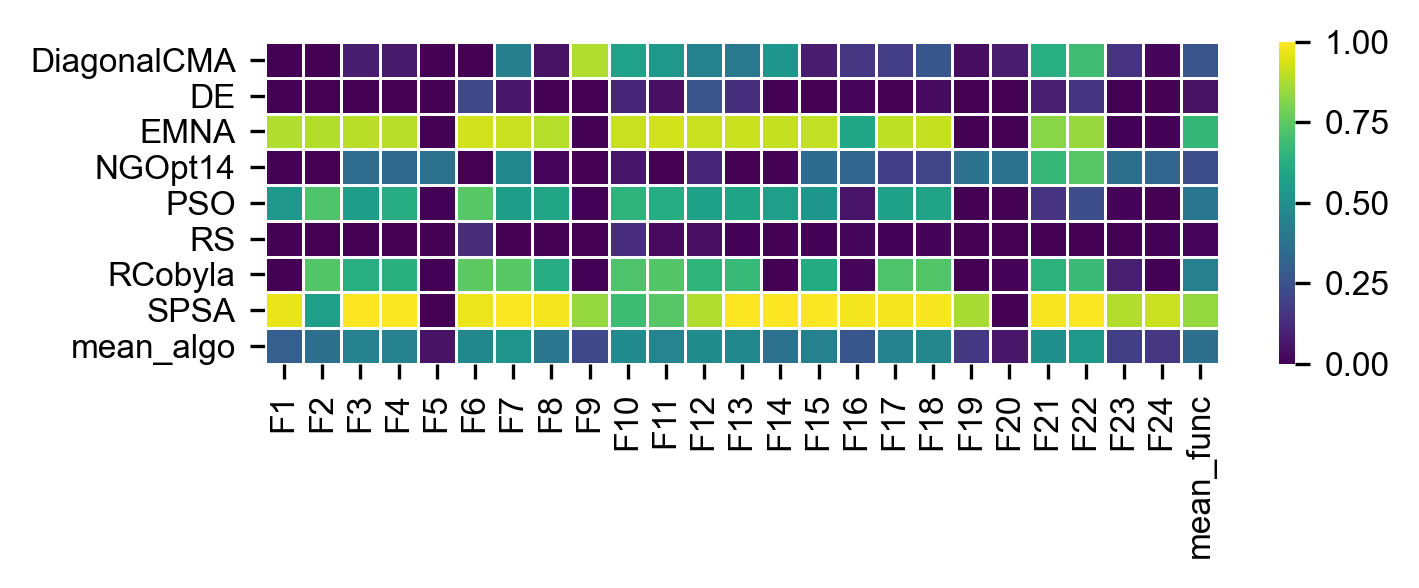}
\includegraphics[width=.5\linewidth,trim=2mm 2mm 17mm 2mm,clip]{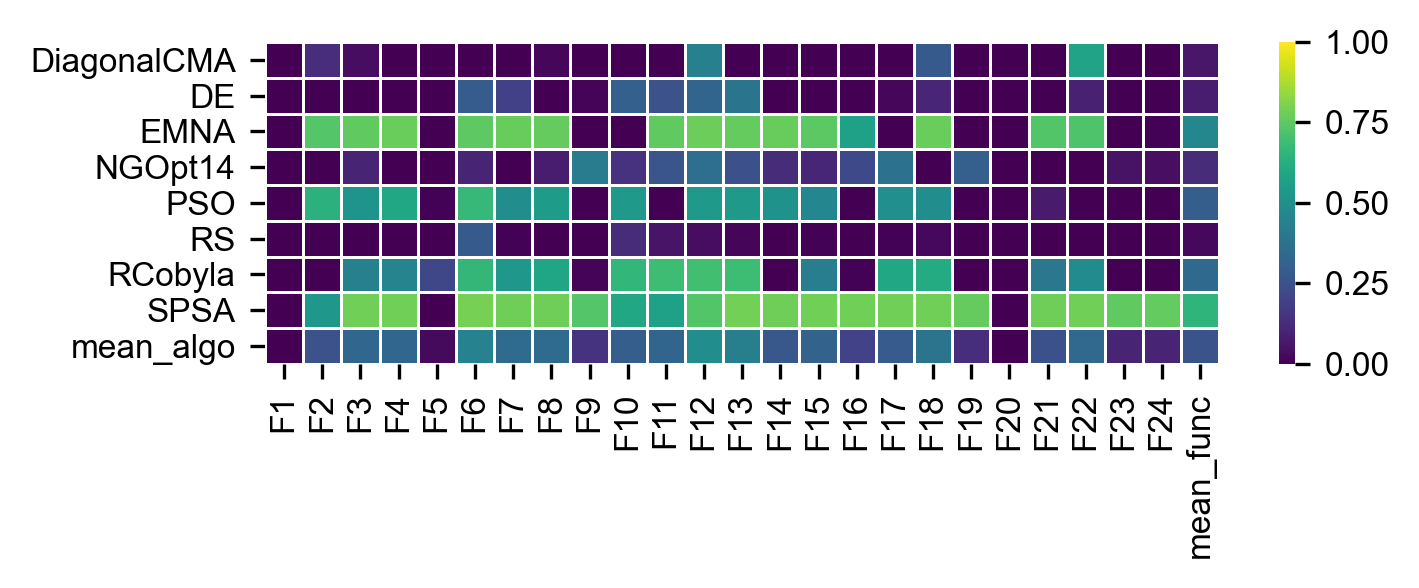}
\includegraphics[width=.49\linewidth,trim=20mm 2mm 2mm 2mm,clip]{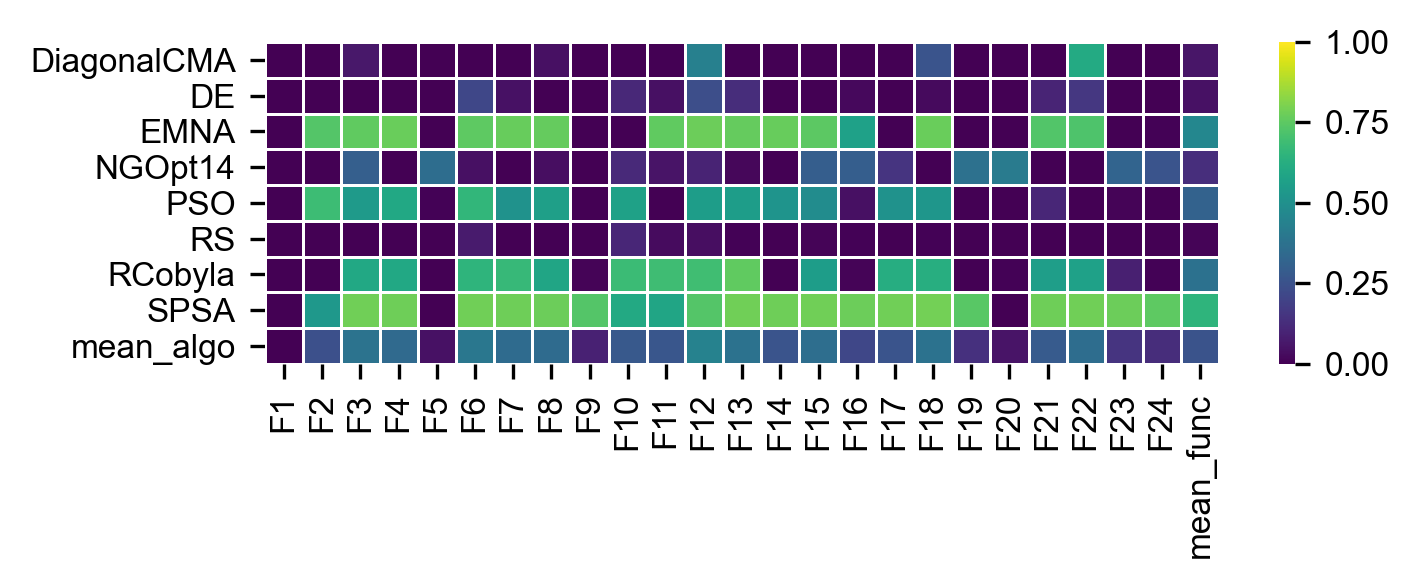}
\caption{Average rejection rate of null hypothesis \textit{algorithm performances are similar across instances}, aggregated over problem instances per function. Left and right column show results for $1\,000$ and $10\,000$ function evaluations, respectively. Top and bottom rows show pairwise and one-vs-all comparisons, respectively. Average values are shown in the last column and row of each figure.}\label{fig:mwu_agg_algo}
\end{figure}

We consider the best function values reached after a fixed-budget of $1\,000$ and $10\,000$ evaluations. 
Since we have 50 runs of each algorithm on each instance, we use a statistical testing procedure to determine whether there are significant differences in performance between instances -- here, we use the Mann-Whitney U (MWU) test with the BH correction method.
In addition to the pairwise testing, we  consider the same procedure in a one-vs-all setting. 
In other words, we repeatedly compare the algorithm performances between the selected instance and the remaining (499) instances. 
The results are visualized in Figure~\ref{fig:mwu_agg_algo}, as fractions of times the test rejects the stated null-hypothesis.

We note that RS indeed seems to be invariant across instances, which is to be expected since we make use of relative performance measure (precision from the optimum) rather than the absolute function values.
Furthermore, with exception of SPSA, all algorithms have stable performance on F5, F19, F20, F23 and F24, which mostly matches the results from Figure~\ref{fig:ks_agg_ela}. 
The fact that SPSA shows differences in performance between these instances, even on F1, shows that this algorithm is not invariant to the transformations used for instance generation. 
This matches with previous observations that SPSA displays clear structural bias~\cite{vermetten2022bias}. 

We would expect several other algorithms, specifically DiagonalCMA and DE, to be invariant to the types of transformation used for the BBOB instance generation.
However, for some problems, e.g. F12 (bent cigar), such assumption does not seem to hold. 
This indicates that for these problems, the instances lead to statistically different performance of these invariant algorithms. 
This might be explainable considering the fact that these algorithms treat the optimization problem as being box-constrained, while the BBOB function transformations make the assumption that the domain is unconstrained~\cite{coco_hansen2021}. 
In addition, while the algorithms might in principle be invariant to rotation and transformation, applying these mechanisms does impact the initialization step, which can have significant impact on algorithm performance~\cite{cmaes_vermetten2022}. 
This is an intended feature of the BBOB suite, since it is claimed that \textit{"If a solver is translation invariant (and hence ignores domain boundaries), this [running on different instances] is equivalent to varying the initial solution"}~\cite{coco_hansen2021}. 
While this is true for unconstrained optimization, it is \textit{not as straightforward} when box-constraints are assumed, as is commonly done when benchmarking on BBOB, since here changing the initialization method might significantly impact algorithm behavior. 

\section{Properties across Instances}\label{sec:result_prop}
For most functions, the general transformation mechanism consists of rotations and translations. 
However, in order to preserve the high-level properties, these transformations are not applied in the same manner for each problem. 
While translation and rotation are indeed the core search space transformations, the order in which they are applied in the chain of transformation which create the final problem can change. 
For simple functions such as the sphere, the transformation is straightforward (a translation only, since rotating a sphere has no impact). 
For other functions, such as the Schaffers10 function (F17), one rotation is applied, followed by an asymmetric function and another rotation, after which the final translation is applied. 
The precise transformations and their ordering is shown in~\cite{bbob_hansen2009_noiseless}. \textit{While these different transformation processes are necessary to preserve the global properties of the problems, their impact on the low-level features of the problem can not always be as easily interpreted}. As a result, the amount of difference between instances on each function is impacted by its associated transformation procedure, which can make some functions much more stable than others.

\begin{figure}[!t]
 \centering
 \includegraphics[width=.24\linewidth,trim=2mm 2mm 2mm 2mm,clip]{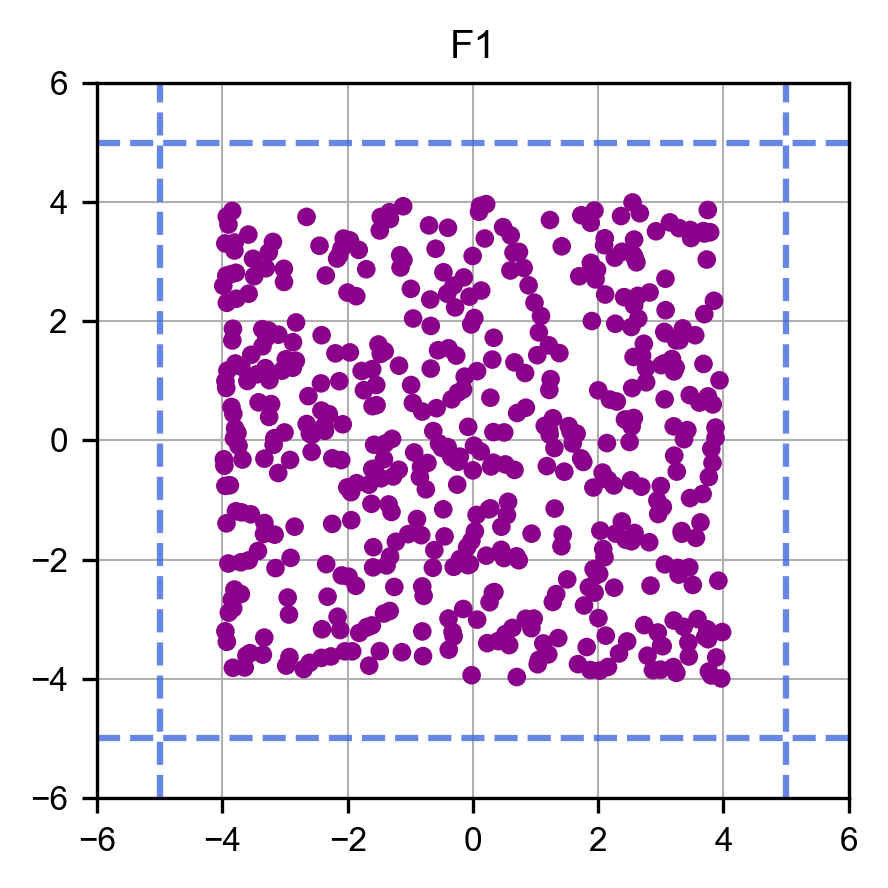}
 \includegraphics[width=.24\linewidth,trim=2mm 2mm 2mm 2mm,clip]{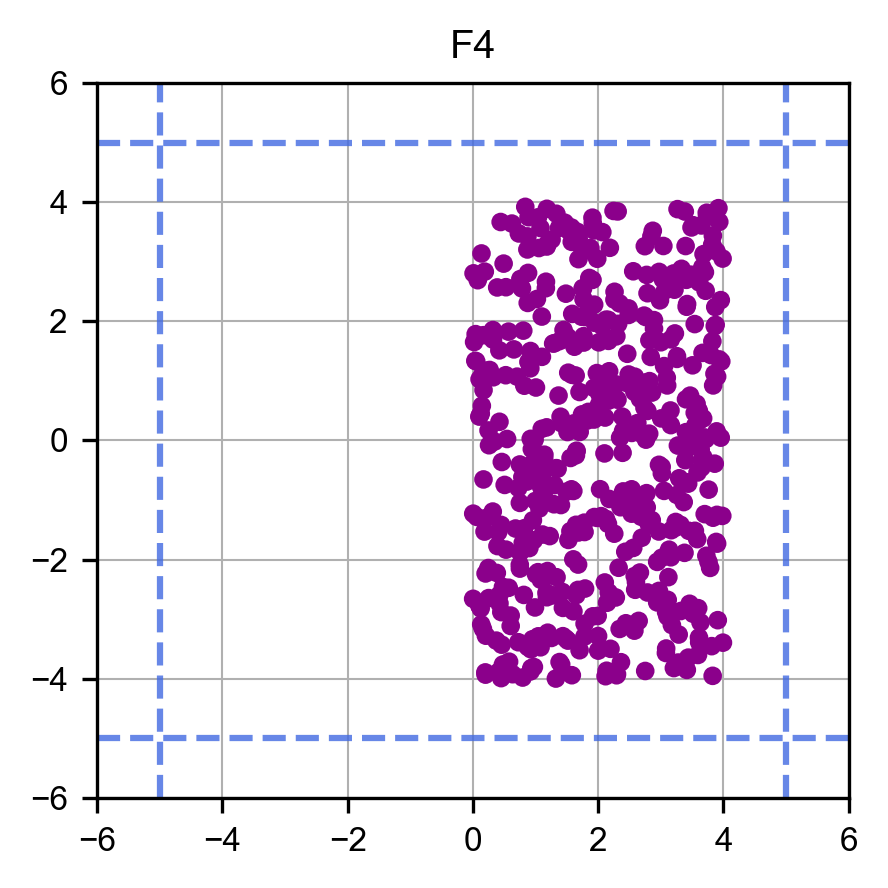}
 \includegraphics[width=.24\linewidth,trim=2mm 2mm 2mm 2mm,clip]{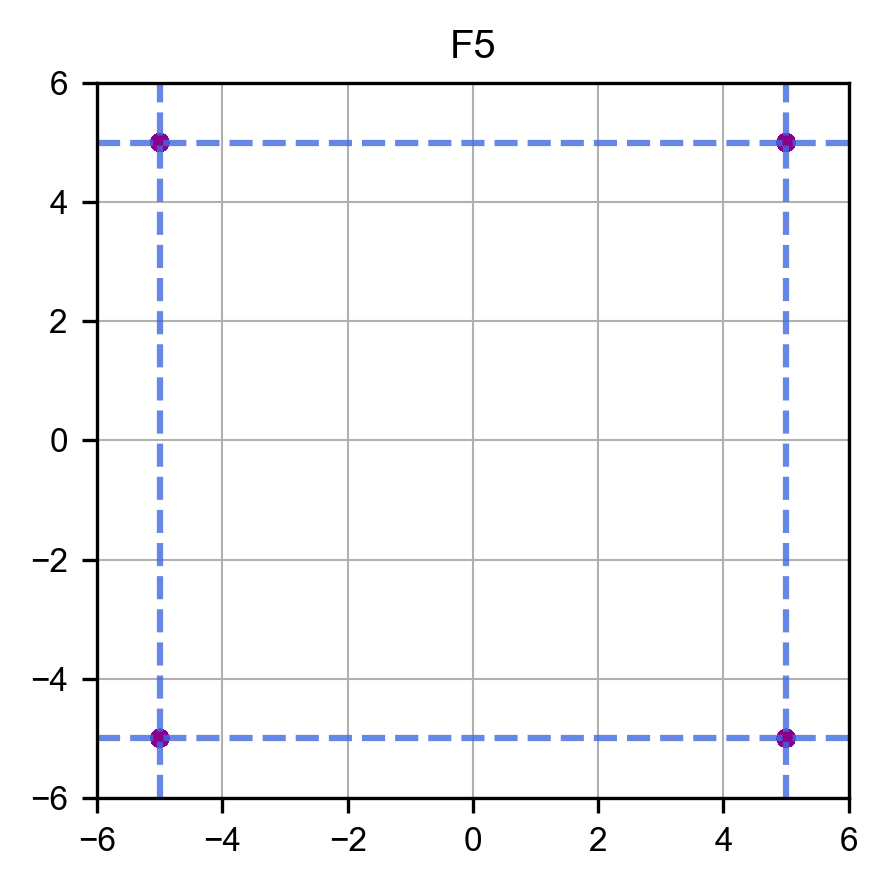}
 \includegraphics[width=.24\linewidth,trim=2mm 2mm 2mm 2mm,clip]{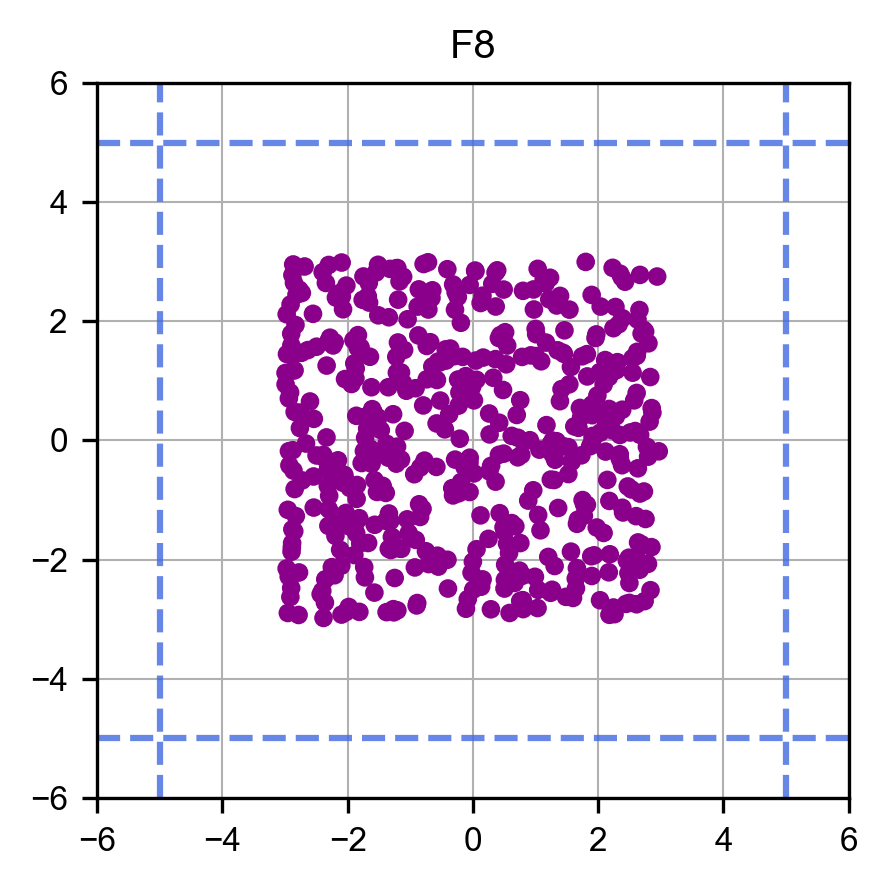}
 \includegraphics[width=.24\linewidth,trim=2mm 2mm 2mm 2mm,clip]{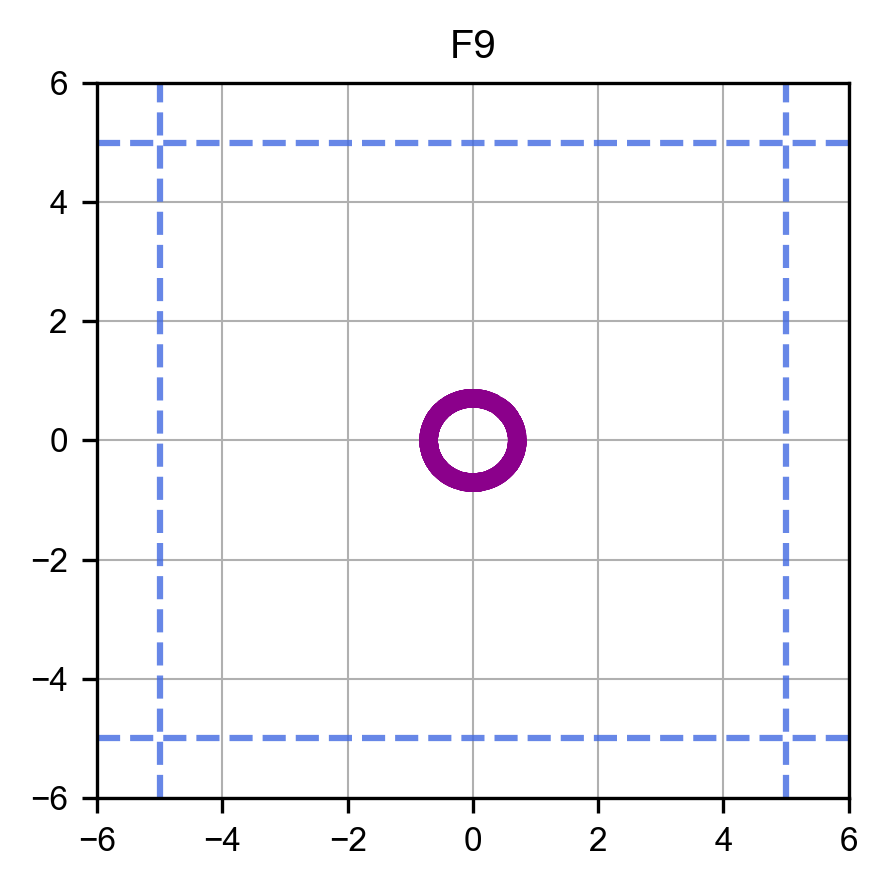}
 \includegraphics[width=.24\linewidth,trim=2mm 2mm 2mm 2mm,clip]{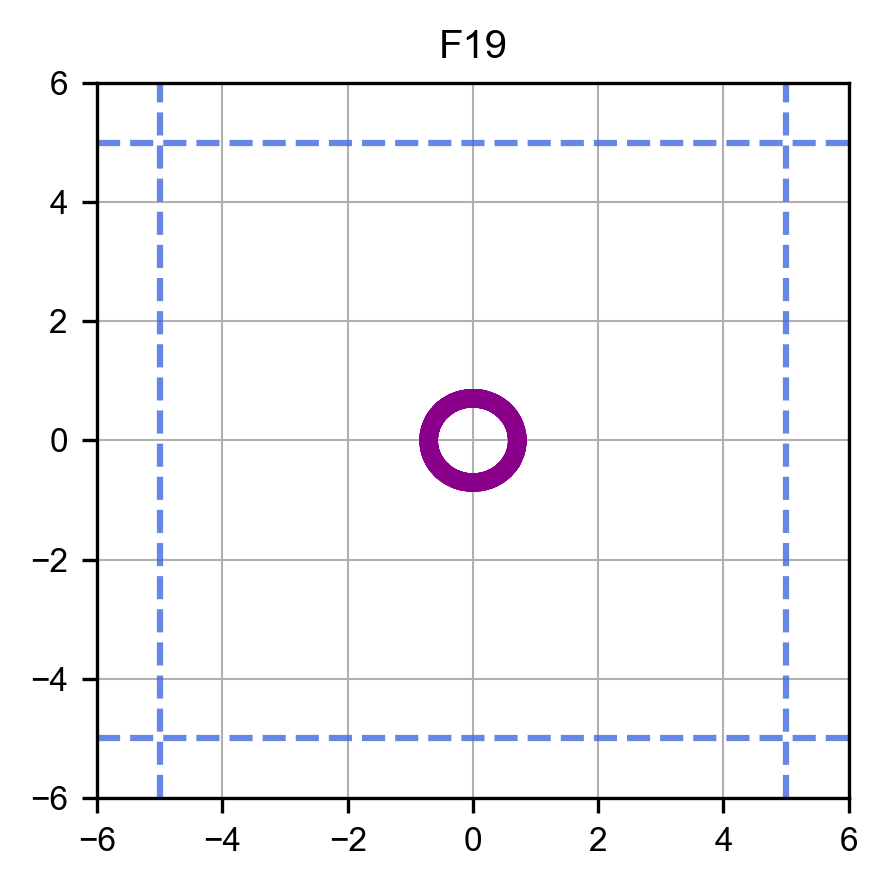}
 \includegraphics[width=.24\linewidth,trim=2mm 2mm 2mm 2mm,clip]{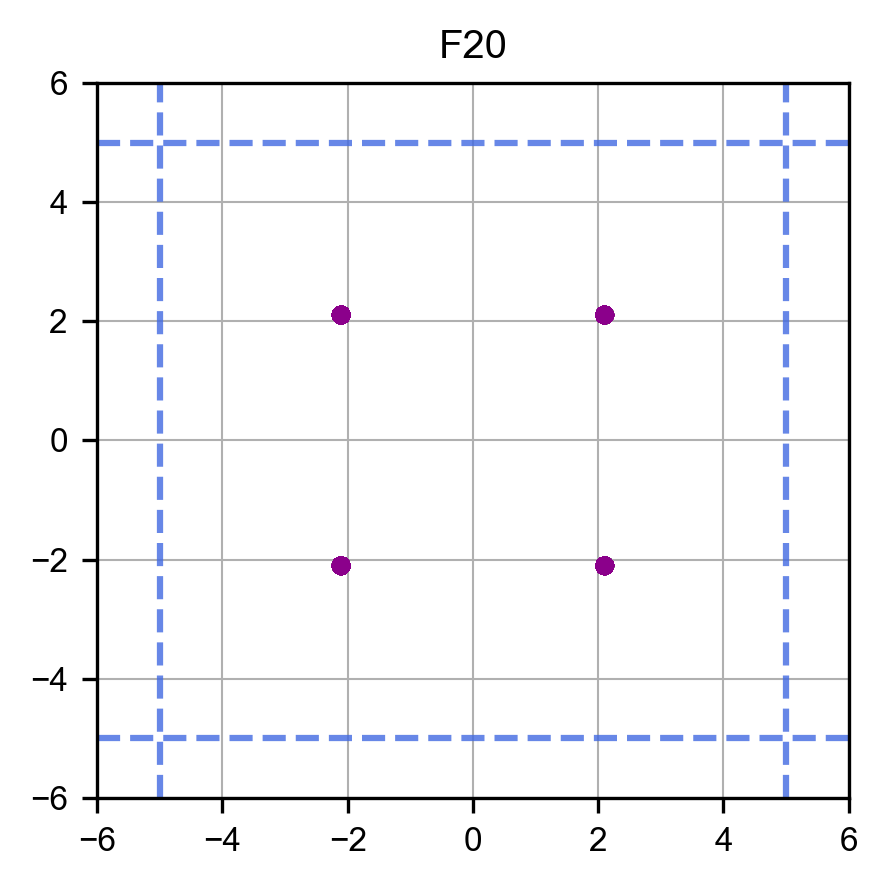}
 \includegraphics[width=.24\linewidth,trim=2mm 2mm 2mm 2mm,clip]{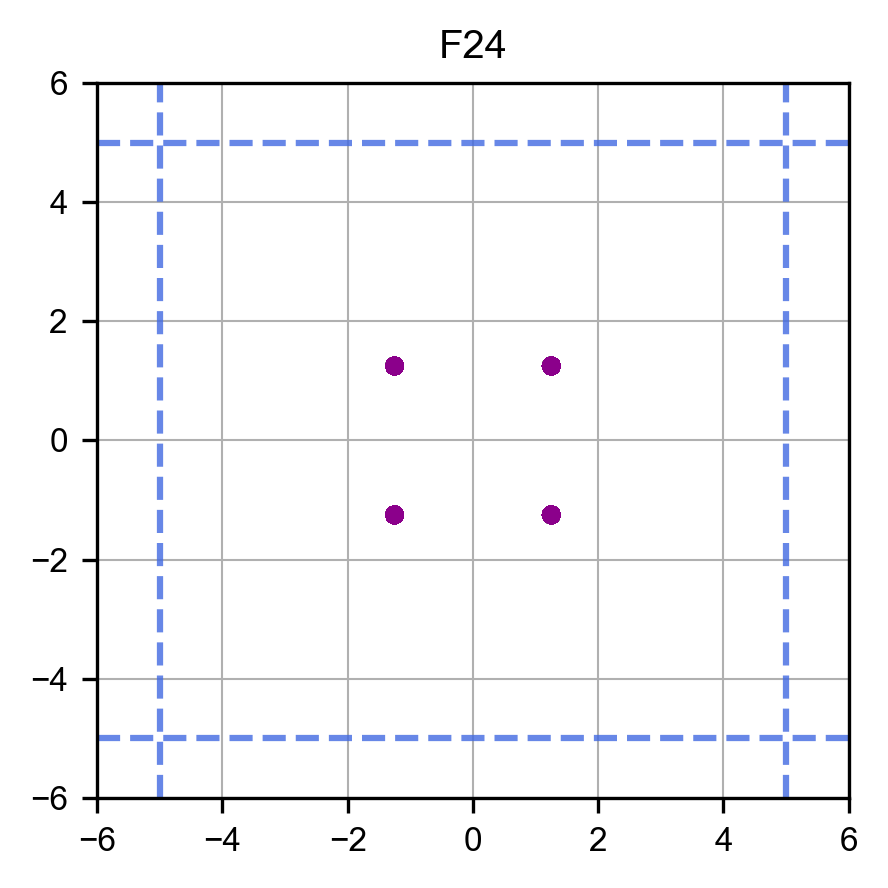}
 \caption{Locations of global optima of $500$ instances for selected BBOB functions in $2d$. Each dot represents the optimum of BBOB instance. The remaining $16$ BBOB functions (not shown here) have a similar distribution pattern as F1. Dashed lines mark the commonly used boundary of search domain $[-5,5]^2$.}
 \label{fig:bbob_xopt}
\end{figure}

One aspect of the instances which is treated differently across problems is the location of the global optimum. By construction, for most BBOB problems, location of this optimum is uniformly sampled in $[-4, 4]^d$. This is achieved by using a translation to this location, since for the default function the optimum is located in $\textbf{0}^d$. 
However, for some other problems, such as the linear slope (F5), a different procedure is used. 
Here, we visualize \textit{true locations of optima across the first 500 instances} of the BBOB functions in $2d$ in Figure~\ref{fig:bbob_xopt}. 
We note that on most functions the situation is equivalent to that of F1, with some exceptions: (i) the asymmetric pattern for F4 (B\"{u}che-Rastrigin) stems from the even coordinates by construction being used in a different way from the odd ones; (ii) on F8 (Rosenbrock), a scaling transformation is applied before the final translation, resulting in the optimum being confined to a smaller space around the optimum; (iii) for the remaining functions (F9, F19, F20, F24), construction of the problem requires a different setup, and as such the optima will be distributed differently.

\begin{figure}[!htbp]
 \centering
 \includegraphics[width=.99\linewidth,trim=7mm 7mm 8mm 6mm,clip]{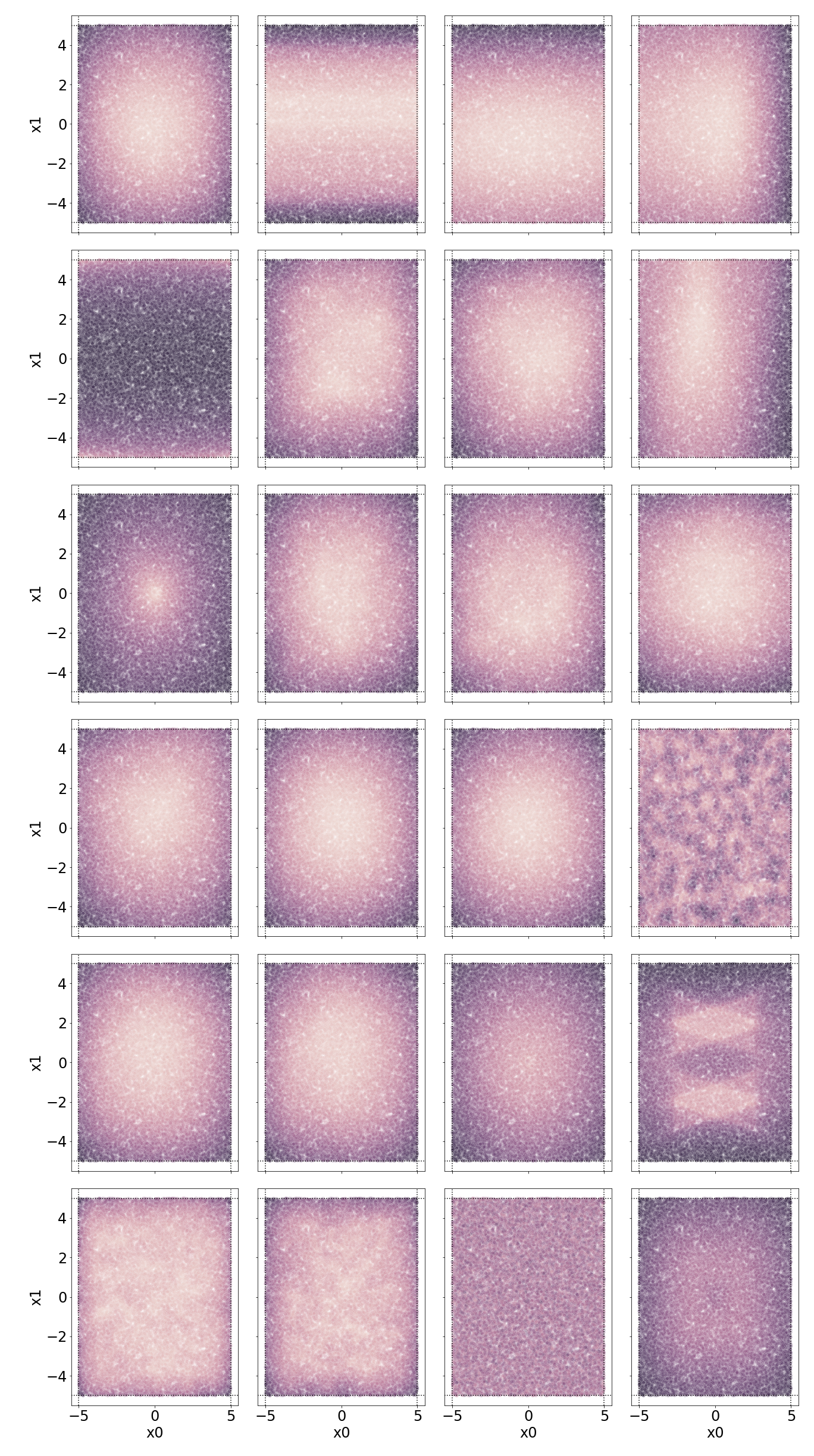}
 \caption{Logarithmic average of relative function value (precision) across the first 500 instances of each BBOB function in $2d$. }
 \label{fig:bbob_avg_fval}
\end{figure}

In addition to considering the location of the optimum, we aggregate the instances together, resulting in an \textit{overview of regions of the space which are on average better performing, across multiple instances}. 
This highlights potential bias in the function definition, see Figure~\ref{fig:bbob_avg_fval}. 
We observe, e.g., that for sphere function (F1), the domain center has a much lower function value on average than the boundaries, which matches our intuition. 
This also indicates that initializing a (reasonably designed) algorithm close to the center might more likely result in good algorithm performance, as we on average directly start with better function values.  
For the BBOB suite overall, we see a \textit{clear skew towards the center of the space}. 
While this is reasonable given the construction of problems (and the underlying implicit assumption that optimization is unconstrained), it potentially hints towards a set of \textit{functions which are not represented in the suite}, namely those which have optima located near the boundaries, or in general give higher fitness to points close to the bounds. 
It is also worth mentioning that, unexpectedly, instance generation on some functions, such as the linear slope (F5) does not lead to equal treatment of dimensions, which results in consistently better regions along the boundary of one dimension only. Such a skew is clearly an artefact of a particular choice of slopes for F5. 
\section{Conclusions and Future Work} \label{sec:conclusion}

In this paper, we investigated differences between instances of the BBOB problems from three main viewpoints. 

Firstly, we see that there are clear differences between functions from the \textit{perspective of low-level ELA features}. 
For some functions, features seem to be mostly similarly distributed, while for others a wide set of features show statistically significant differences in their distributions. 
While this effect seemingly lessens with increasing dimension, there are still many functions where instance differences are clearly present. 
This seems to indicate that care should be taken relying on the low-level features to represent instances, e.g. in an algorithm selection context, as the choice of which instances to include in training or testing could potentially lead to different results. 

Secondly, from the \textit{perspective of algorithm performance}, we found that only random search is close to being fully invariant with regard to changing instances. 
While algorithms such as CMA-ES are typically thought of as rotation invariant, there still are cases where differences in performance between instances of the same function can be observed.
This might be related to the impact the transformations have on the effectiveness of initialization, as was hypothesised earlier~\cite{cmaes_vermetten2022}, or it might be related to the fact that we considered the problems to be box-constrained, which seemingly invalidates some of the assumptions made when the transformation methods were designed. 

Lastly, the final viewpoint in which we observe differences between instances is the \textit{perspective of global function properties}, specifically the location of the optimum in the search space and, as a consequence, the average values of function within the domain across instances. 
While for most problems, the former is confirmed to be uniform at random in $[-4,4]^d$, several problems do not follow this pattern because their problem formulation seemingly requires a different transformation mechanism. This raises the question on whether \textit{additional transformation mechanisms} might need to be introduced to better balance the available instances. 

While this work aimed to provide some insights into the properties of the instances as used by the BBOB suite, there are still many \textit{unanswered questions}. 
In particular, we have not yet established a clear link between differences observed from the ELA-perspective and the algorithm performance on those instances.
Investigating this relation in more detail has the potential to reveal some links between the landscape properties and algorithmic behavior. 
In addition, one could study the impact of the transformation methods in a more isolated setting, to gain a better understanding into the impact this has on each of the ELA features. 
Such an understanding would allow us to use transformations in combination with other benchmarking suites, which might not originally have been designed with instance generation in mind. 

Finally, as mentioned multiple times throughout this paper, there is a certain level of ambiguity in the `box-constrained vs. unconstrained' nature of BBOB, which clearly impacts algorithm design choices. 
We believe further elaboration on this question is required. 

\subsubsection{Acknowledgements} 
The contribution of this paper was written as part of the joint project newAIDE under the consortium leadership of BMW AG with the partners Altair Engineering GmbH, divis intelligent solutions GmbH, MSC Software GmbH, Technical University of Munich, TWT GmbH. The project is supported by the Federal Ministry for Economic Affairs and Climate Action (BMWK) on the basis of a decision by the German Bundestag. This work was performed using the ALICE compute resources provided by Leiden University.

\bibliographystyle{splncs04}
\bibliography{main.bib}

\end{document}